\documentclass[runningheads]{llncs}
\usepackage[table]{xcolor}
\usepackage{adjustbox}
\usepackage{graphicx}
\usepackage{dirtytalk}
\usepackage{enumitem}
\usepackage{colortbl}
\usepackage{hhline}
\usepackage{multirow}
\begin{document}
\title{Modeling Trust in Human-Robot Interaction: A Survey}
%
%
\author{Zahra Rezaei Khavas\inst{1}\orcidID{0000-0002-5268-0197}  \and
S. Reza Ahmadzadeh\inst{2}\orcidID{0000-0002-6651-8684}
\and
Paul Robinette\inst{1}\orcidID{0000-0001-8066-156X}}
\authorrunning{Z. Rezaei et al.}
%
\institute{Electrical and Computer Engineering, University of Massachusetts Lowell, Lowell, MA 01854, USA \email{Zahra\_RezaeiKhavas@student.uml.edu, paul\_robinette@uml.edu} \and Computer Science, University of Massachusetts Lowell, Lowell, MA 01854, USA \email{reza\_ahmadzadeh@uml.edu}}

%
\maketitle              
\begin{abstract}
 
As the autonomy and capabilities of robotic systems increase, they are expected to play the role of teammates rather than tools and interact with human collaborators in a more realistic manner, creating a more human-like relationship. Given the impact of trust observed in human-robot interaction (HRI), appropriate trust in robotic collaborators is one of the leading factors influencing the performance of human-robot interaction. Team performance can be diminished if people do not trust robots appropriately by disusing or misusing them based on limited experience. Therefore, trust in HRI needs to be calibrated properly, rather than maximized, to let the formation of an appropriate level of trust in human collaborators. For trust calibration in HRI, trust needs to be modeled first. There are many reviews on factors affecting trust in HRI \cite{hancock2011meta}, however, as there are no reviews concentrated on different trust models, in this paper, we review different techniques and methods for trust modeling in HRI. We also present a list of potential directions for further research and some challenges that need to be addressed in future work on human-robot trust modeling.
\end{abstract}
\section{Introduction}

Trust is one of the essential factors in the development of constructive relationships, including relationships among human individuals and automation. We can mention trust as an overarching concern that affects the effectiveness of a system, especially in terms of safety, performance, and use rate \cite {lee2004trust}. Having this concern in mind, trust has become a critical element in the design and development of automated systems \cite{martelaro2016tell}. Autonomous systems are being designed and developed with increased levels of independence and decision-making capabilities, and these capacities will be efficient in uncertain situations \cite{schaefer2016meta}.

Human-robot trust is an important branch of Human-Robot Interaction (HRI), which has recently gained increasing attention among scholars in many disciplines, such as Computer Engineering \cite{robinette2016overtrust}, Psychology \cite{hancock2011meta}, Computer Science \cite{desai2012modeling} and Mechanical Engineering \cite{sadrfaridpour2016modeling}. Trust is a significant factor that needs to be taken into consideration when robots are going to work as teammates in human-robot teams \cite{groom2007can}, used as autonomous agents \cite{selkowitz2015effects}, or where robots are going to be used in a complex and dangerous situations \cite{ososky2014determinants,robinette2016overtrust}. In many cases, trust is the main factor determining how much a robotic agent would be accepted and used by the human \cite {parasuraman1997humans}.

Weak partnerships resulting from inappropriate or non-calibrated trust between humans and robots might cause misuse or disuse of a robotic agent. Misuse refers to the failures that occur due to a user’s over-trust of the robotic agent (e.g., accepting all the solutions and results presented by the robot without questioning). In contrast, disuse refers to the failures that occur due to human under-trust to the robotic agent (e.g., rejecting the capabilities of a robotic agent) \cite{lee2004trust}. To prevent human operators’ misuse and disuse of robots, trust needs to be calibrated. We need to model trust to generate a measure of it and to enable proper trust calibration  \cite{ososky2013building}.

There are many reviews available in HRI which are mostly concentrated on factors affecting trust \cite{hancock2011meta,schaefer2016meta}, but none of them is concentrated on modeling trust and different trust models in HRI. This document's purpose is to review different studies concentrated on trust modeling in HRI. This review starts by defining trust, followed by summarizing factors affecting trust, reviewing studies focused on modeling trust, and a summary of input and output elements to the trust models. We conclude with a discussion of the shortcomings and challenges of trust modeling and future avenues of research. 

\section{Definition of Trust in HRI}

According to psychologists, trust is a mental state of a human \cite{atoyan2006trust}. Numerous researchers have extensively explored the notion of trust for decades. Trust is not limited to just interpersonal interactions. It underlies different forms of interactions, such as banks' interaction with customers, governments with citizens, employers with employees, etc., \cite{wagner2011recognizing}. Therefore, we can say trust can affect human-robot interaction, as it can affect a human user or collaborator's willingness to assign tasks, share information, cooperate, provide support, accept results, and interact with a robot~\cite{freedy2007measurement}.

As trust is one of the necessities for building a successful human-robot interaction, we need to create methodologies to model, measure, and calibrate trust. The first step for modeling trust is a clear definition of trust. However, despite the broad efforts and number of studies concentrating on trust, there is little consensus on a single definition, as the definition of trust is heavily dependent on the context in which trust is being discussed \cite{cameron2015framing}. This is highlighted, for instance, in \cite{oleson2011antecedents}, where trust is defined as \say{\textit{the reliance by one agent that actions prejudicial to the well-being of that agent will not be undertaken by influential others}}. Around each application and domain that robots are used, trust needs to be defined, measured and explored explicitly. For example, in high-risk robotic applications such as emergency evacuation robots \cite{robinette2016overtrust}, the definition of trust might differ substantially from low-risk robotic applications such as border tracking robots \cite{xu2015optimo}.
One of the most thorough definitions of trust, which is deployed by many other studies concentrated on human-robot trust, is by Lee and See \cite{lee2004trust}. They define trust from the perspective of automation. This definition was generated by reviewing many other studies concentrated on defining trust and was complementary for many other works. They define trust as: \say{\textit{the attitude that an agent will help achieve an individual’s goals in a situation characterized by uncertainty and vulnerability}}. This definition of trust is accepted and used by many studies on trust in HRI. Wagner et al. \cite{wagner2011recognizing} also provided a comprehensive definition for trust: \say{\textit{a belief, held by the trustor, that the trustee will act in a manner that mitigates the trustor’s risk in a situation in which the trustor has put its outcomes at risk}}. They also provided a model for determining if an interaction demands trust or not. All these definitions have one thing in common, that is: ``whether robot’s actions and behaviors correspond to human’s interest or not?'' To address this concern in each robotic application trust needs to be modeled based on human interests in that domain.

\section{Factors Affecting Trust}


Studies on factors affecting trust in HRI can be considered an extension to the studies on factors affecting human-automation trust. In \cite{lee2004trust}, the authors review the factors affecting trust in humans and generalize these factors to factors affecting trust in automation. Many other studies also review and analyze factors affecting trust in human-automation Interaction (HAI) \cite{hoff2015trust,schaefer2016meta}. However, robots differ from other forms of automation in many cases, such as mobility, embodiment, and unfamiliarity to the general public. Therefore, factors affecting trust in HRI need to be investigated separately.

Hancock et al. \cite{hancock2011meta} provides a meta-analysis of factors affecting trust in HRI and classifies these factors in three categories each consist of two subcategories: 1. Human-related factors (i.e., including ability-based factors, characteristics); 2. Robot-related factors (i.e., including performance-based factors, attribute-based factors); and 3. Environmental factors (i.e., including team collaboration, tasking). In this study, we review and classify recent studies on factors affecting trust with a similar classification basis as in \cite{hancock2011meta} with some updates in categories and subcategories. We classify factors affecting trust in HRI into three categories: 1. Robot-related factors (including robot-performance, robot-appearance, and robot-behaviors), 2. Human-related factors and 3. Task and Environment-related factors. Table \ref{tab:my-table} shows our classification of factors affecting trust. Most of these factors are similar to the ones that are mentioned in Hancock et al. \cite{hancock2011meta}. However, some factors are excluded as recent studies paid less attentions to them and some new factors that are stressed more in recent studies are added to this table.Some factors are also classified under different categories here in this work compared to \cite{hancock2011meta}.

 \begin{enumerate}

\item \textbf{Robot-Related Factors:}
Robot-related factors have the greatest effect on the trust in HRI \cite{hancock2011meta}. This justifies the great number of studies concentrated on robot-related factors. We classify robot-related factors under three subcategories: 

\begin{enumerate}[label=(\alph*)]

\item  Performance-related factors: These factors determine the quality of an operation performed by the robot from the human operator's point of view. Of these factors we can mention \emph{reliability}, \emph{faulty behavior},  \emph{frequency of fault occurrence} \cite{desai2012effects}, \emph{timing of error} \cite{lucas2018getting},\emph{transparency}, \emph{feedback} \cite{salem2015would,natarajan2020effects}, \emph{level of situation awareness} \cite{boyce2015effects}, \emph{false alarms} \cite{yang2017evaluating}, and \emph{level of autonomy} \cite{lazanyi2017dispositional}.

\item Behavior-related factors: Advancements in robotic systems in recent years have caused people to consider them more like teammates than tools. Increased autonomy of robots caused an increase in the perceived intelligence of robots by humans. These advancements altered the form of HRI to a more naturalistic interaction \cite{ososky2013building}. Approaching a more human-like interaction with robots cause people to consider the intention of a robot's behavior. Some general behaviors of the robot such as \emph{likeability} (e.g., gaze behaviors and greeting) \cite{mumm2011human}, \emph{proximity} (e.g., physical and physiological proximity) \cite{mumm2011human,obaid2016stop,walters2011long}, \emph{engagement} \cite{robinette2013building}, \emph{confess to the reliability} \cite{wang2016trust,natarajan2020effects} and \emph{harmony of robot personality with the task} (e.g., introverted robotic security guard and an extroverted robotic nurse) \cite{tay2014stereotypes} can affect formation and maintenance of trust. There are also some specific behaviors of robots that can affect trust repair after failure or wrong and misleading action such as \emph{apology}, \emph{making excuses} or \emph{explanations} and \emph{dialogues} \cite{robinette2015timing,lucas2018getting,strohkorb2018ripple,sebo2019don,natarajan2020effects}.

\item Appearance-related factors: People consider the personality of a robot based on the robot's appearance and behavior during interactions. Some features in robot appearance such as \emph{anthropomorphism} \cite{natarajan2020effects}, \emph{gender} \cite{tay2014stereotypes}, \emph{harmony of task with robot's appearance} \cite{robinette2013building}, and \emph{similarity with human collaborator} \cite{you2018human} affect trust in HRI.
\end{enumerate}

\item\textbf{Human-Related Factors:}
Although, according to \cite{hancock2011meta}, human-related factors have the least effect on trust, they are important in the formation and fluctuation of trust during human-robot interaction. Many studies investigated the effect of these factors on trust. Of these factors, individual's \emph{gender},  \emph{subjective feeling} toward robots~\cite{obaid2016stop}, \emph{initial expectations} of people toward automation~\cite{yang2017evaluating},\emph{ previous experience} of the individuals with robots~ \cite{walters2011long}, \emph{culture} \cite{li2010cross} and also the human's \emph{understanding of the system} ~\cite{ososky2013building} are some of the human-related factors that are addressed in different studies. 

\item\textbf{Task and Environment-Related Factors:}
Based on Hancock et al. \cite{hancock2011meta}, task and environment-related factors have the second greatest effect on human-robot trust. Many factors related to the the task type and location such as \emph{rationality} and \emph{revocability} of the tasks~\cite{salem2015would}, \emph{risk} and \emph{human-safety}~\cite{robinette2016overtrust}, \emph{workload}~\cite{desai2012effects}, \emph{nature of the task} (e.g., nurse or security robot) \cite{tay2014stereotypes}, \emph{task duration} \cite{walters2011long,yang2017evaluating}, \emph{in-group membership} \cite{groom2007can,rau2009effects}, \emph{physical presence of robot} \cite{bainbridge2011benefits,natarajan2020effects} and \emph{task site}~\cite{lazanyi2017dispositional} are thoroughly investigated by researchers in HRI.

\end{enumerate}





\begin{table}
\caption{Factors affecting development of trust in human-robot interaction}
\label{tab:my-table}
\resizebox{\textwidth}{!}{%
\centering

\begin{tabular}{|l|l|l|}
\hline
\rowcolor[HTML]{9B9B9B} 
\multicolumn{2}{|c|}{\cellcolor[HTML]{9B9B9B}1. Robot-Related}                                                                                                         & 3. Task \& Environment-Related                                                                                        \\ \hline
\multicolumn{1}{|c|}{\cellcolor[HTML]{C0C0C0}(a) Performance-Related}                     & \cellcolor[HTML]{C0C0C0}(c) Appearance-Related                             & Nature of task \cite{tay2014stereotypes,lazanyi2017dispositional}                                                                                                       \\ \hline
\begin{tabular}[c]{@{}l@{}}Dependability, reliability and \\  error \cite{desai2012effects,desai2013impact,salem2015would,robinette2016overtrust,natarajan2020effects,lucas2018getting} \end{tabular}        & Similarity with operator  \cite{you2018human}                                                  & \begin{tabular}[c]{@{}l@{}}Physical presence of robot\\ in task site \cite{bainbridge2011benefits,natarajan2020effects}\end{tabular}                                     \\ \hline
Level of \cite{lazanyi2017dispositional,goodrich2001experiments}                                                                        & Gender \cite{you2018human,tay2014stereotypes}                                                                    & In-group membership    \cite{rau2009effects,groom2007can}                                                                                               \\ \hline
\begin{tabular}[c]{@{}l@{}}Situation awareness, feedback \\ and Transparency \cite{desai2013impact,natarajan2020effects,boyce2015effects,wang2016trust}\end{tabular} & \begin{tabular}[c]{@{}l@{}}Harmony of appearance with \\ task \cite{robinette2013building}\end{tabular} & Duration of interaction \cite{yang2017evaluating,walters2011long}                                                                                              \\ \hline
\multicolumn{1}{|c|}{\cellcolor[HTML]{C0C0C0}(b) Behavior-Related}                        & Anthropomorphism  \cite{natarajan2020effects}                                                          & Task site \cite{lazanyi2017dispositional}                                                                                                            \\ \hline
Dialogues \cite{lucas2018getting}                                                                                & \multicolumn{1}{c|}{\cellcolor[HTML]{9B9B9B}2. Human-Related}              & Revocability \cite{salem2015would}                                                                                                         \\ \hline
\cellcolor[HTML]{FFFFFF}Proximity \cite{mumm2011human,obaid2016stop,walters2011long}                                                        & \cellcolor[HTML]{FFFFFF}Personality \cite{salem2015would}                                       & Rationality \cite{salem2015would,robinette2016overtrust,bainbridge2011benefits}                                                                                                          \\ \hline
Likeability and friendliness \cite{mumm2011human}                                                             & Culture \cite{li2010cross}                                                                   & Risk \cite{robinette2016overtrust,lazanyi2017dispositional,you2018human}                                                                                                                 \\ \hline
\begin{tabular}[c]{@{}l@{}}Personality (harmony with\\  task) \cite{tay2014stereotypes}\end{tabular}                & \begin{tabular}[c]{@{}l@{}}Understanding of the\\  system \cite{ososky2013building}\end{tabular}     &                                                                                                                       \\ \cline{1-2}
Confess to reliability \cite{wang2016trust,natarajan2020effects}                                                                   & Demographics \cite{obaid2016stop,m2011measuring}                                                               & \multirow{-2}{*}{\begin{tabular}[c]{@{}l@{}}Workload, complexity and \\required level of multi tasking  \cite{desai2012effects}\end{tabular}} \\ \hline
Apology for failure \cite{strohkorb2018ripple,sebo2019don,natarajan2020effects}                                                                      & Subjective feeling \cite{obaid2016stop,yang2017evaluating}                                                        &                                                                                                                       \\ \hline
Engagement  \cite{robinette2013building}                                                                               & Experience with robots \cite{m2011measuring}                                                    &                                                                                                                       \\ \hline
\end{tabular}

}

\end{table}

\section{Current Research on Trust Modeling}

Trust in HRI has a lot in common with trust in HAI,which has been studied at length. Muir et al. \cite{muir1994trust} found the available definitions for trust between humans inconsistent with the nature of HAI based on the multidimensional construct of trust. She defined a trust models for HAI which was based on model of human expectation of automation proposed by Barber et al. \cite{barber1983logic}. Lee and Moray \cite{lee1992trust} built upon Muir's strategy for modeling trust, which was identifying independent variables that influence trust, and introduced another model for trust. Later, other researchers have modeled the operator's trust in automation, considering more factors affecting trust \cite{itoh2000mathematical,farrell2000connectionist,lee2004trust}. These models were finally classified into five groups \cite{moray1999laboratory}: regression-based models \cite{muir1994trust,lee1992trust}, time series models \cite{lee1994trust}, qualitative models \cite{riley1996operator}, argument based probabilistic models\cite{cohen1998trust}, and neural net models \cite{farrell2000connectionist}.

There are many similarities between trust in HRI and trust in HAI. However, most of the models generated for modeling trust in HAI are inconsistent with the needs of HRI. According to Desai \cite{desai2009creating} \say{\textit{these models do not consider some factors that appear while working with robots such as situational awareness, usability of the interface, physical presence of robots (co-located with human or remote-located), limitations and complexities of the operating environment, workload, task difficulty, etc. which influence HRI considerably}}. Desai et al. \cite{desai2009creating} introduced a schematic of a model considering some factors affecting human-robot trust in conjugate with factors affecting human-automation trust. Yagoda et al. \cite{yagoda2012you} introduced one of the very first models for human-robot trust based on the different dimensions of a human-robot interaction task and validity assessment of each of these directions by subject matter experts (SMEs) in the HRI. Desai et al.  \cite{desai2012modeling} also was one of the pioneers in modeling trust in HRI. They generated a more detailed model for trust in human and autonomous robot tele-operation. This model used the Area Under Trust Curve (AUTC) measure to account for an individual’s entire interactive experience with the robot. 

According to several studies, there is a strong correlation between the level of trust in human-robot teammates with the performance of the robotic agent’s work, and it also impacts their interaction quality \cite{lee2004trust,desai2012modeling}. According to Xu \cite{xu2016maintaining}, high levels of trust among human-robot teammates often demonstrate great synergy, in which matched decision-making capabilities of the human member in the team complements the exhaustive controlling and executing capabilities of the robotic agent. In contrast, a low level of trust among human-robot teammates might cause a human to refuse to delegate tasks to the robotic agent or sometimes decide to disable the robotic agent \cite{xu2016maintaining}. Since there is often a high correlation among trust and performance of the work in human-robot collaboration, trust has been modeled based on the performance in many studies  \cite{xu2012trust,sadrfaridpour2016modeling}. Most of these performance-based trust models are used as a feedback loop to adjust actual performance of the robot to the human's expectation of its performance  to convince a human to act trusting toward robot. There are also some trust models based on the performance of robot operations, which are not aimed to modify the performance of collaboration. For example, in \cite{pippin2014trust} a performance-based trust model for multi-robot tasks is designed to detect robotic agents that are not reliable. Then, less reliable agents are assigned to do less critical tasks or sometimes disregarded while assigning tasks. On the other hand, the human-robot collaboration's performance can also be modeled based on trust \cite{freedy2007measurement,gao2013modeling}. These models are used to modify the robot’s trust-related behavior to manage the overall performance of the collaboration.

A prevalent class of human-robot collaborations is supervisory collaboration in which the human plays the role of supervisor and the robot plays the role of worker. The supervisor delegates tasks to the worker and oversees the performance of the operation. The supervisor also has the authority to take control of the robot when the robot is doing something wrong. The model of trust for supervisory collaboration presented in \cite{xu2012trust} is based on trust in between-human collaboration. It generates a quantity showing the compatibility of robot performance with the human expectation, enabling the robot to modify its performance to fulfill human expectations and improve trust. Later this trust model was improved \cite{xu2016towards}, and more factors affecting trust in supervisory collaboration such as failure rate in the autonomous agent and the rate of supervisor intervention were involved in designing the trust model. The Online Probabilistic Trust Inference Model (OPTIMo) \cite{xu2015optimo} is another model of trust in the supervisory collaboration introduced by the same research group. This model formulates Bayesian beliefs over the human’s trust status based on the robot's performance on the task over time to generate a real-time estimate of the human’s trust.


When trust can be modeled and measured in a real-time manner in human-robot collaboration, it can help the robot repair trust whenever the human starts under-trusting the robot \cite{xu2015optimo}. A real-time model of trust (trust-POMDP) for human-robot peer-to-peer collaboration is introduced in \cite{chen2018planning}, which integrates measured trust in the robot’s decision-making. The trust-POMDP model closes the loop between measured trust by the real-time trust model and robot decision-making process to maximize collaboration performance. This model grants a robot the ability to influence human trust systematically to reduce and increase trust in over-reliance and under-reliance situations, respectively.


Hancock et al. \cite{hancock2011meta} provided a meta-analysis of a great number of factors affecting trust in HRI, and quantitative measurement for the effect of human-based, robot-based, and environment-based factors on trust. Later they developed a model of human-robot team trust based on their findings in the meta-analysis \cite{hancock2011meta}. In addition to the effect of three classes of factors affecting trust introduced in the prior work, they considered the effects of training and design implications on the final model of trust \cite{sanders2011model}.

Some recent studies in trust modeling use subjective trust measurement techniques in companions with objective trust measurement techniques. These techniques are deployed in both HAI and HRI for increasing the accuracy and robustness of trust measurements. There are some studies in human-automation trust, human-computer interaction, and human trust in artificial intelligence that use psycho-physiological measurements for trust modeling \cite{ajenaghughrure2019predictive,gulati2017modelling,gulati2019design}. Khalid et al. \cite{khalid2016exploring} also introduce a trust model in HRI, which uses facial expressions, voice features, and extracted heart rate features in combination with the self-reported trust of humans to model trust. This trust model classifies the trust level into one of the low, natural, and high trust levels using a Neuro-fuzzy trust classifier.

\section{Trust Model: Inputs and Outputs}
Trust models formulate the effect of factors on the formation and variation of trust in robots. In fact, trust models use factors affecting trust to estimate trust. Since these factors vary in different domains and environments, input factors to the trust models vary based on the application domain. For example, Robinette et al. \cite{robinette2014modeling} models trust in emergency evacuation based on the situational risk (e.g., amount of danger perceived by the human in the environment around him) and agent risk (e.g., agent's behavior and appearance) to model perceived trust by the human and the human's decision to trust the robot's guidance or not. In contrast, \cite{xu2012trust} proposes a trust model for a supervisory collaboration and formulates trust as a function of the robot's success and failure in performing the task. The output of this trust model is closing the loop between human trust and robot function by adjusting the robot's action to improve the collaboration efficiency. Finally, \cite{yagoda2012you} proposes a more general trust model based on team configuration, task, system, context, and team process to scale trust for trust measurement. 

Many of the studies on modeling trust in HRI consider the performance of collaboration as one of the main input elements for their model \cite{xu2015optimo,gao2013modeling,sadrfaridpour2016modeling}. Most of these models consider the effect of performance in conjugate with some other factors. For instance, the OPTIMo probabilistic trust model \cite{xu2015optimo} uses rates of robot's failures, and human interventions in conjugate with task performance as inputs to the model to estimate the human's degree of trust in a robotic teammate. Meanwhile, \cite{gao2013modeling} uses the operator's perception of system capabilities, past experience, and training to assess initial trust. Trust gets updated in a loop based on system performance, cognitive workload, and frequency of changes from tele-operation to autonomous operation. This trust model's output is a measure of gain and loss of trust and the impact of these changes of trust on collaboration performance. Sadrfaridpour et al. \cite{sadrfaridpour2016modeling} models trust based on human performance (i.e., muscles fatigue and dynamics of recovery), robot performance (i.e., speed of robot doing the specific task), workload and human expectation of task performance. The output of this model is feedback to the robot to adjust its performance according to operator desires.

\section{Conclusion and Future Work}
Most of the existing trust models in HRI are developed for a specific form of human-robot interaction or a specific type of robotic agent. For example, in \cite{robinette2014modeling}, a model of trust is specified for evacuation robots; in \cite{desai2012effects} the trust model is specified on robots with shared control, and in \cite{xu2015optimo} the generated model is usable for supervisory collaboration. As each of these trust models belongs to a specific domain, they can not be compared with each other, and there is no scale to evaluate their accuracy. A general model of trust is needed in HRI to evaluate the accuracy of other trust models. Such a model can be applied to any robotic domain and even can be deployed by newly emerged robotic areas and eliminate the need for new trust models for these areas. 

Trust is a subject of interest for research in many other fields such as psychology, sociology, and even physiology. In these fields, other indicators of trust are used for trust measurement. For example, some studies use physiological indicators, such as oxytocin-related measures \cite{uvnas2005oxytocin,ferreira2018relationship,johnson2011trust}, and objective measures, such as trust games that assess actual investment behavior \cite{chang2010seeing,keri2009sharing}. These trust assessment methods can be used in HRI to develop a trust model that is independent of the countless parameters that affect trust.

Many studies examined the effect of robot failure on trust \cite{robinette2016overtrust,desai2012effects,salem2015would,natarajan2020effects}. However, there are limited studies focused on modeling the effect of robot failure on trust. While Desai et al. \cite{desai2012modeling} investigated the effect of robot's reduced performance and timing of performance reduction on the operator's trust in a tele-operation human-robot interaction, the effect of trust violation needs to be investigated more deeply. Different forms of trust violation in different robotic domains need to be explored. A variety of factors affecting trust-loss and trust-repair after trust violation, such as task type, risk, robot type, robot behavior, etc. need to be taken into account in modeling the effect of robot failure on trust. Modeling fluctuations of trust in a human after a robot failure would help estimate the time required for human trust to tend to a steady-state and formulate timing of trust repair after trust violation.

Both modeling trust and modeling the effects of failure on trust need to be explored in a more general context. Since there are many different factors affecting trust, it will be challenging to develop a conclusive trust model that incorporates all of these factors. Therefore, future research into trust modeling in HRI needs to be more focused on developing general models of trust, which are based on measures other than the countless factors affecting trust. Such models would not be affected by the emergence of new factors affecting trust in existing or new robotic domains and eliminate the need to develop new trust models for new domains of the ever-evolving world of robotics.

\bibliographystyle{splncs04}
\bibliography{trust-modeling}

\end{document}